\documentclass{article}
\usepackage{spconf,amsmath,graphicx}
\usepackage{multirow}
\usepackage{graphicx}
\usepackage{algorithm}
\usepackage{algorithmic}
\usepackage{subfigure}
\usepackage{amsmath}
\usepackage{amssymb}
\usepackage{booktabs}
\usepackage[citecolor=red]{hyperref}
\usepackage{color}
\usepackage{multirow}
\usepackage{bbding}
\usepackage{xcolor}

\title{ Class Relevance Learning for Out-of-Distribution Detection}
\name{Butian Xiong \textsuperscript{\dag}, Liguang Zhou \textsuperscript{\dag}, Tin Lun Lam \textsuperscript{*} , Yangsheng Xu \thanks{\textsuperscript{*} Corresponding Author: Tin Lun Lam (tllam@cuhk.edu.cn)} 
\thanks{\dag \, Authors contributed equally to this work.}
\thanks{ This work was partly supported by the National Natural Science Foundation of China under Grant 62073274, the Shenzhen Science and Technology Program under the Grant JCYJ20220818103000001, and the Shenzhen Institute of Artificial Intelligence and Robotics for Society under Grant AC01202101103.}
}

\address{Chinese University of Hong Kong, Shenzhen, China, \\ 
	Shenzhen Institute of Artificial Intelligence and Robotics for Society}

\begin{document}
\maketitle

\begin{abstract}
Image classification plays a pivotal role across diverse applications, yet challenges persist when models are deployed in real-world scenarios. Notably, these models falter in detecting unfamiliar classes that were not incorporated during classifier training, a formidable hurdle for safe and effective real-world model deployment, commonly known as out-of-distribution (OOD) detection. While existing techniques, like max logits, aim to leverage logits for OOD identification, they often disregard the intricate interclass relationships that underlie effective detection. This paper presents an innovative class relevance learning method tailored for OOD detection. Our method establishes a comprehensive class relevance learning framework, strategically harnessing interclass relationships within the OOD pipeline. This framework significantly augments OOD detection capabilities. Extensive experimentation on diverse datasets, encompassing generic image classification datasets (Near OOD and Far OOD datasets), demonstrates the superiority of our method over state-of-the-art alternatives for OOD detection.

\end{abstract}


\section{INTRODUCTION}
Image classification is a well-studied task in computer vision and robotics. It aims to recognize the image with a trained classifier. Various methods have been developed to better represent image representations, with varying levels of success. These methods can be divided into different categories, including the development of stronger network architectures \cite{zhou2022feature}, semantic-based enhancer \cite{zhou2021borm}\cite{miao2021object}, and multi-modality learning \cite{zhou2022attentional}.

Despite these successes in image classification task, many classifiers fail to generalize to an open set setting, wherein an image from an unknown class is mistakenly classified as a known class. For instance, the parking lot is erroneously classified as a garage due to the presence of a car in the space; the difference is that the car is parked on the road as opposed to a garage. Similarly, the restaurant mistakenly classified as a dining room due to the presence of many dining tables in the room - a feature that is more commonly seen in restaurants compared to a home dining room.

There are various methods emerged for OOD detection \cite{chen2023detecting,chen2023batch,kahya2023mcrood,bukhsh2023out,zhou202osm}. Recent studies have sought to address the challenge of out-of-distribution detection by introducing an intra-class splitting method \cite{smith2022openscenevlad}. This technique aims to create atypical subsets of the known classes that can be used to model the unknown abnormal classes \cite{schlachter2019open}. However, this method tends to increase the risk of falsely rejecting known classes as unknown classes.

\begin{figure}[t]
	\begin{center}
		\includegraphics[width=8cm,height=5cm]{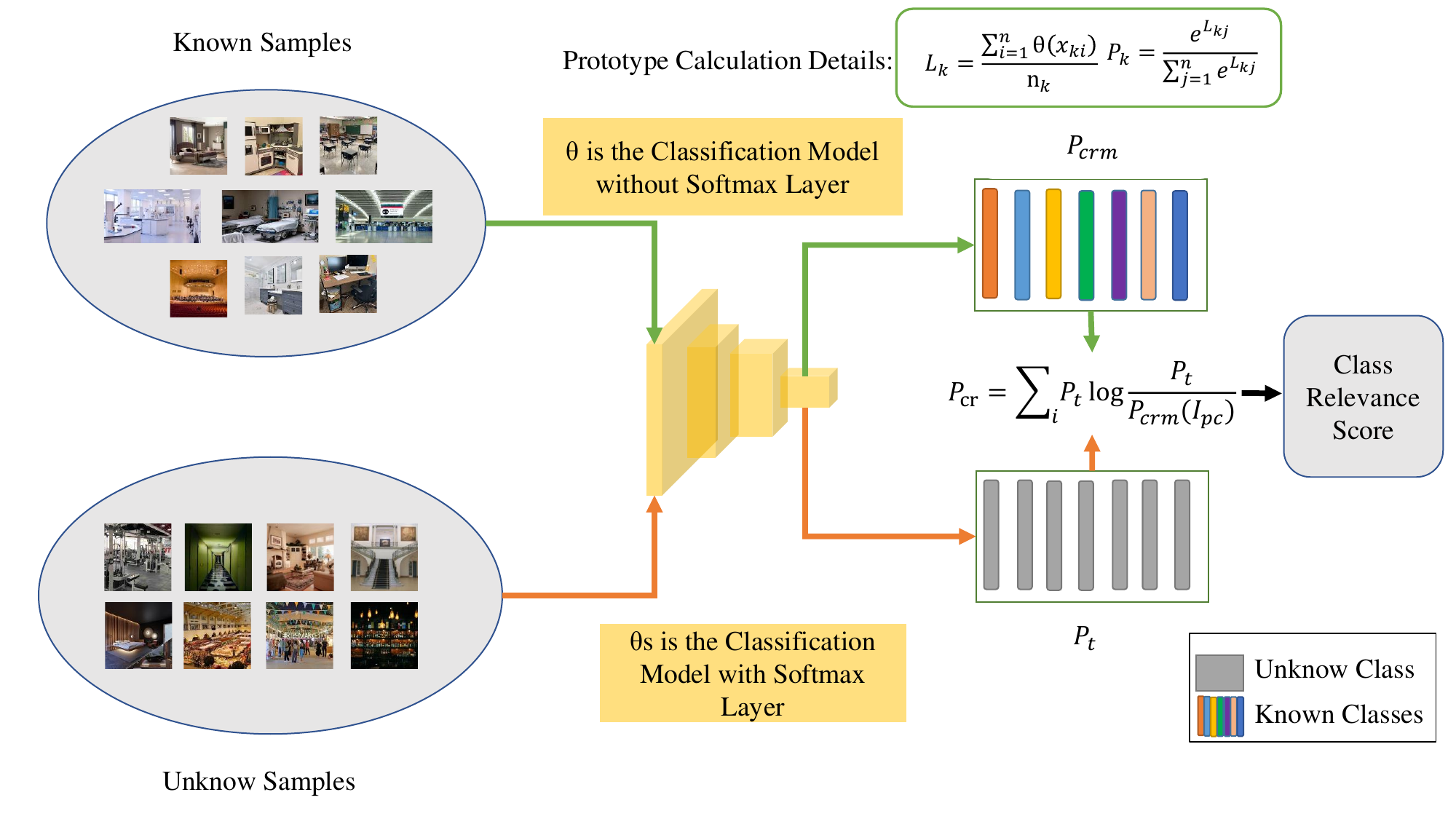}
	\end{center}\vspace*{-10pt}\caption{Proposed class relevance learning framework for measuring the class relevance score of a test sample to the constructed class relevance matrix of training dataset.}
	\label{fig:method}
\end{figure}

In out-of-distribution (OOD) detection, many methods have been proposed to identify samples from a distribution different from the training dataset. Common strategies include using the maximum softmax probability \cite{hendrycks2016baseline} or max logits \cite{hendrycks2019anomalyseg}. It is assumed that if a sample is correctly classified, its maximum value could be exploited for OOD detection. However, these methods neglect the relevance of class relationships, which is also important for analyzing OOD samples. Standardized max logits \cite{jung2021standardized}, show their findings that max logits on the range of max logits to the predicated classes. This phenomenon causes unexpected semantic classes predicated as a certain class. Therefore, standardization technology is applied to address this problem.  These methods show the great potential for using the output of the model for OOD detection.

However, these methods have only considered the logits/softmax probability of the class itself when judging out-of-distribution (OOD) samples, neglecting the relevance of class relationships which is essential for OOD judgment. To address this, we propose a class relevance learning framework to learn prototypes of each class into two levels. At the logits level, the maximum class logits is utilized, while at the same time, the class relevance prototype is developed to capture the relationship between different classes. This framework takes into account the relevance of class relationships, thus allowing for a more comprehensive judgment of OOD samples. 

Our main contributions are summarized as follows: 1)  We propose a simple yet effective post-processing method, namely class relevance learning to statistically compute the class relevance matrix for in-distribution (ID) classes. 2) Different from previous methods, including MSP, max logits, and standardized max logits, that merely exploit the logits/softmax probability on the class level, we first take the class relevance matrix into consideration. 3) Extensive experimental results on diverse image classification datasets verify the superior performance of the proposed method for OOD detection.

\section{Methodology}
\label{sec:method}

\subsection{Problem Statement}
Out-of-distribution detection is a learning problem wherein a model is trained with an ID dataset of labeled images, denoted as $D_{ID}$, to recognize various categories, represented as $C_{i}=\{1,2,...,n_i\}$, where $n_i$ is the number of classes in the training set. During the test,  the model  is tested on both ID and OOD datasets, with some categories not present in the training set, denoted as $D_{OOD}$. The number of classes in the test set is denoted as $C_o=\{1,2,...,n_i,..,n_o\}$, where $n_o$ is the total number of classes in the test set, and the difference between $n_o$ and $n_i$ indicates the number of OOD classes. The model should be able to recognize the known samples in the ID classes while detecting the unknown samples in the OOD classes.

\subsection{Class Relevance Learning}
Traditional OOD detection algorithms only have a fixed prototype that merely uses the network outputs, such as logits/softmax probability, which is limited and does not exploit class relationships. To address this, we propose a novel class relevance learning framework to statistically describe the class relevance among each class in the $D_{ID}$ dataset after the training phase. Then, the class relevance matrix of the training dataset is statistically established. 

As indicated in Fig.\ref{fig:method}, the model is firstly trained on the $D_{ID}$ with known samples. Then, the parameters of the model are fixed. After that, the classification model without the softmax layer, denoted as $\theta$, is obtained. The $\theta$ is utilized to obtain the class prototype by averaging the output logits of each class in the known samples as $L_k$, where $n_k$ is the number of samples of $k-{th}$ class in the training dataset, and $x_{ki}$ is i-th images in the $k-th$ classes.

\begin{align}
	{L_k} & = \frac{\sum_{i=1}^{n_k} \theta(x_{ki})}{n_k} \in  R ^ {n\times 1} \\
	{P_k} &  =  \sigma(L_k) = \frac{e^{L_{kj}}}{\sum_{j=1}^{n}e^{L_{kj}}} \in  R ^ {n\times 1} \\
	{P_{crm}} &  = \{P_1, .., P_i, .., P_K\} \in R ^ {n \times n} 
\end{align}

Then, the softmax version of average logits $L_k$ is calculated as $P_k$, where the  $L_{kj}$ denotes $j^{th}$ column of prototype logit of class k. Therefore, the output $P_k$ is the prototype probability of $k-th$ class.

We iteratively calculates the prototype of each classes with this process. Finally, the class relevance matrix, denoted as $P_{crm}$, is constructed through this process, where each row represents the prototype probability of a certain class. 

After obtaining the class relevance matrix, we can calculate the distance between a test sample and class relevance matrix. First, the softmax of a test sample can be otained and defined as $P_{t}$, where the $\theta_s$ is the classification model with the softmax layer and $z$ is a test image. $I_{pc}$ means the pseudo class index that best matches the input sample $z$. The reason why it is pseudo class is that it may not be the true class of the input sample. Then, to check out where is the closest class to test sample, index of pseudo class $I_{pc}$ of a test sample is introduced by taking the Argmax function on $P_t$, where the class number of most likely ID classes is identified. Hence, we can estimate the distance between a test sample to the prototype of most relevant class for the ID dataset by looking at the class relevance matrix $P_{crm}$.

\begin{equation}
	\begin{split}
		P_{t}&=\theta_s(z) \in R^{n \times 1} \\
		I_{pc} &= argmax(\theta_s(z)) \\
		P_{cr}  &= \sum_{i} P_{t} \log \frac{P_{t}}{P_{crm}(I_{pc})}
	\end{split}
	\label{eq:dz}
\end{equation}

The class relevance score, denoted as $P_{cr}$, quantize the distance between a test sample to its pseudo class prototype. $P_{cr}$ will serve as a measurement for the system to choose which sample is OOD and which is ID. If $P_{cr}$ is large, the distance between a test sample to its pesudo class prototype is large. In other words, this sample is very likely to be an OOD sample. Instead, if $P_{cr}$ is small, it means the relevance of the a test sample its pseudo class prototype is small, which indicates the test sample is very likely to be an ID sample.

\begin{equation}
	P_{cf} = -max(\theta(z))*\alpha - \frac{1}{P_{cr}} * \beta
\end{equation}
In addition to introducing the class relevance score, we also preserves the maximum logits as a complementary score, which can be regarded as the class score. The final sample OOD score, denoted as $P_{cf}$, quantifies the degree to which a sample may be considered an OOD sample, where $\alpha$ is 5, $\beta$ is 0.5 by default. These parameters controls the influence of max logits and class relevance score for OOD detection. Lower values of $P_{cf}$ correspond to higher likelihoods of an ID sample, whereas higher values of $P_{cf}$ correspond to greater likelihoods of an OOD sample.

\section{Experiments and Results}
\label{sec:exp}

\subsection{Experimental Setup}
\textbf{Datasets}
To examine the generalization ability of the proposed method for OOD detection, we used the generic image classification datasets, including the CIFAR10 as an ID dataset. Two benchmarks were used for OOD detection: Near-OOD and Far-OOD \cite{yang2022openood}. In the Near-OOD setting, ID samples are prevented from being wrongly introduced into OOD sets. The Near-OOD dataset contains samples that are closer to the ID dataset, and are therefore more difficult to distinguish from it. The Far-OOD dataset, on the other hand, contains samples that are more easily distinguished from the ID dataset, as they are significantly different. The CIFAR100 \cite{krizhevsky2009learning} and TinyImageNet \cite{le2015tiny} datasets were used as Near-OOD datasets. The MINIST \cite{deng2012mnist}, SVHN \cite{netzer2011reading}, Texture \cite{kylberg2011kylberg}, and Places365 \cite{zhou2017places} datasets are used as far OOD datasets after 1,305 images were removed due to semantic overlap. The detailed test set split for OOD can be found in \cite{yang2022openood}.

\textbf{Evaluation Metrics} 
The performance of OOD detection are evaluated using the following metrics: false positive rate at 95\% true positive rate (FPR95) and area under the receiver operating characteristic curve (AUROC).

\subsection{Experimental Results}
\begin{table*}[t]
	\caption{Comparison of our method with other state-of-the-art approaches using various datasets. The ID dataset is CIFAR10, while the OOD dataset consists of two settings: Near-OOD and Far-OOD. The backbone network employed is ResNet18.}
	\label{exp:Classification_OOD}
	\begin{tabular}{lcccccccc}
		\hline \multirow{2}{*}{ { OOD Dataset }} & \multicolumn{2}{c}{{ ODIN} \cite{liang2017enhancing} } & \multicolumn{2}{c}{ { DICE } \cite{sun2022dice}} & \multicolumn{2}{c}{{ SHE} \cite{zhang2022out} }  & \multicolumn{2}{c}{ CRL (Ours)} \\
		& { FPR95 } $\downarrow$ & { AUROC } $\uparrow$ & { FPR95 } $\downarrow$ & { AUROC } $\uparrow$ & { FPR95 } $\downarrow$ & { AUROC } $\uparrow$ & { FPR95 } $\downarrow$ & { AUROC } $\uparrow$ \\
		\hline { CIFAR100 } & 74.80 & 83.21  & 71.41 & 77.17 & 79.34 & 80.67 & 46.47 & 88.77 \\
		{ TinyImageNet } & 72.98 & 84.83 & 66.21 & 78.83 & 76.54 & 83.34 &  38.22 & 90.63 \\ \hline
		{ Near-OOD } & 73.89 & 84.02 & 68.81 & 78.00 & 77.94 & 82.01 & \textbf{42.34} & \textbf{89.70} \\ \hline
		{ MNIST } & 41.27 & 92.47 & 45.62 & 82.01 & 70.96 & 83.87 &  29.10 & 91.87   \\
		{ SVHN } & 69.01 & 85.56 & 30.79 & 91.06 & 66.70 & 85.33 & 27.72 & 91.96 \\
		{ Texture } & 57.08 & 89.22 & 68.34 & 79.28 & 81.56 & 82.73 & 28.40 &  92.07  \\
		{ Places365 } & 74.02 & 84.58 & 77.79 & 74.02 & 80.18 & 81.34 & 41.84 &  90.00 \\ \hline
		{ Far-OOD } & 60.34 & 87.96 & 55.64 & 81.59 & 74.85 & 83.32 & \textbf{31.77} & \textbf{91.48} \\ \hline
	\end{tabular}
\end{table*}

In this section, we present the results of proposed CRL method in comparison to several state-of-the-art techniques on various datasets using ResNet18 as the backbone network. The goal is to assess the performance of CRL, in the context of OOD detection. Table~\ref{exp:Classification_OOD} presents a detailed comparison of CRL (Ours), with three other prominent techniques in the field: ODIN \cite{liang2017enhancing}, DICE \cite{sun2022dice}, and SHE \cite{zhang2022out}. The table showcases the performance of these methods on various datasets and provides insights into their effectiveness in OOD detection.

\begin{table}[h]
	\small
	\centering
	\caption{Ablation study. The comparison between the Maxlogits and CRL is displayed.}
	\label{tab:abl}
	\begin{tabular}{l|ccccc}
		\hline		
		\multirow{2}{*}{OOD Dataset}	& \multicolumn{2}{c}{{Maxlogits \cite{hendrycks2019anomalyseg}	 }} & \multicolumn{2}{c}{{CRL (Ours)}} \\ \cline{2-5}
		& FPR95$\downarrow$ & AUROC$\uparrow$ &  FPR95$\downarrow$ & AUROC$\uparrow$ \\ \hline
		CIFAR100 & 64.97 & 86.60 &  46.47 & 88.77 \\
		TinyImageNet & 55.08 & 89.07 & 38.22 & 90.63 \\ \hline
		Near-OOD  	&  60.02       &  87.84  &   \textbf{42.34} & \textbf{89.70}    \\  \hline
		MINIST & 43.23 & 90.66 & 29.10 &  91.87  \\
		SVHN & 43.68 & 90.37 &  27.72 & 91.96 \\
		Texture & 45.67 & 90.30 & 28.40 &   92.07 \\
		Places365 & 61.93 & 88.08 & 41.84 &  90.00 \\ \hline
		Far-OOD   	& 48.63   &  89.85		& \textbf{31.77} & \textbf{91.48}  \\ \hline
	\end{tabular}
\end{table}

In the Near-OOD setting, CRL achieves an outstanding FPR95 of 42.34\%, significantly outperforming ODIN (73.89\%), DICE (68.81\%), and SHE (77.94\%). Moreover, CRL exhibits an impressive AUROC score of 89.70\%, surpassing the competitors by a substantial margin. In the challenging Far-OOD scenario, CRL maintains its superiority with an FPR95 of 31.77\%, substantially lower than ODIN (60.34\%), DICE (55.64\%), and SHE (74.85\%), accompanied by a remarkable AUROC score of 91.48\%.

In conclusion, our proposed method, CRL, demonstrates remarkable performance in OOD detection across a diverse range of datasets. It consistently outperforms existing state-of-the-art techniques, as evidenced by the substantial reduction in FPR95 and the high AUROC scores. This suggests that CRL  has the potential to significantly enhance the reliability and robustness of AI systems, making it a valuable contribution to the field of OOD detection.

\textbf{Ablation Study}
We present the results of an ablation study designed to assess the contributions of our method. We compare CRL, against baseline method: Maxlogits \cite{hendrycks2019anomalyseg}. The evaluation is performed on two settings of the out-of-distribution (OOD) dataset: Near-OOD and Far-OOD. The primary goal is to analyze the impact of class relevance information components on the performance of OOD detection. Table~\ref{tab:abl} presents the results of the ablation study. 
In the Far-OOD setting,  Maxlogits exhibits slightly lower FPR95 at 48.63\% and an AUROC of 89.85\%. Once again, CRL stands out by achieving the lowest FPR95 of 31.77\% and the highest AUROC of 91.48\%. This highlights the robustness and effectiveness of CRL in identifying Far-OOD samples. In the more challenging Near-OOD setting, Maxlogits performs worse with an FPR95 of 60.02\% and an AUROC of 87.84\%. Notably, our proposed CRL outperforms Maxlogits with an impressive FPR95 of 42.34\% and an AUROC of 89.70\%. These results demonstrate the superiority of CRL in effectively detecting samples that are near the in-domain distribution.
\begin{table}[]
	\small
	\centering
	\caption{Ablation study. The selection of hyper parameters.}
	\label{tab:abl_hyper}
	\begin{tabular}{cc|cc}
		\hline		
		\multicolumn{2}{c}{Parameters}	& {{Near-OOD}} & {{Far-OOD}} \\ \hline
		$\alpha$ & $\beta$ & AUROC$\uparrow$  & AUROC$\uparrow$\\ \hline
		1.0 & 0.5 &	 89.15 &  90.92	\\
		2.0 & 0.5 &  89.27 &  91.04	\\
		5.0 & 0.5 &  {89.42}  & {91.20} \\  
		5.0 & 0.7 &  89.47 & 91.25 \\  
		5.0 & 1.0 &  89.52 & 91.30 \\  	
		5.0 & 3.0 &  89.66 & 91.43 \\			
		5.0 & 5.0 &  \textbf{89.70} & \textbf{91.48} \\	\hline
	\end{tabular}
\end{table}

In Table \ref{tab:abl_hyper}, we delve into the impact of two key hyperparameters, $\alpha$ and $\beta$, on the performance of our model in OOD detection tasks. Focusing on the case with a fixed $\alpha = 5.0$, we observe that varying $\beta$ has a discernible effect on both Near-OOD and Far-OOD AUROC scores. Notably, as $\beta$ increases from 0.5 to 5.0, both Near-OOD and Far-OOD AUROC scores exhibit a consistent upward trend, culminating in peak performances of 89.70 for Near-OOD and 91.48 for Far-OOD at $\beta = 5.0$. These findings emphasize the crucial role of hyperparameter fine-tuning, particularly with respect to $\beta$, in enhancing the robustness of our model for OOD detection tasks.

\begin{figure}[t]
	\begin{center}
		\subfigure{\includegraphics[width=4cm]{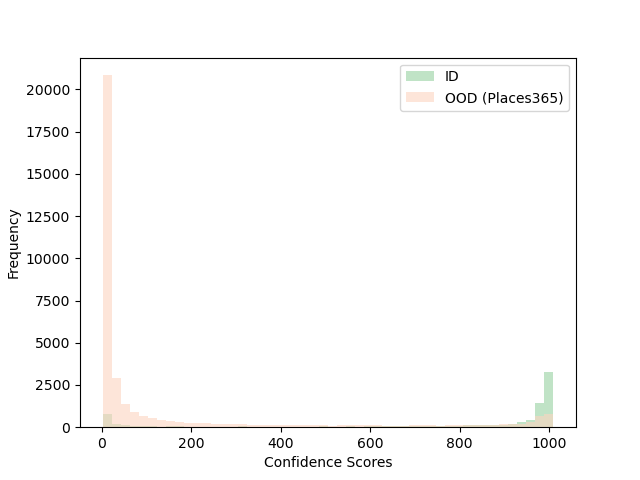}}
		\subfigure{\includegraphics[width=4cm]{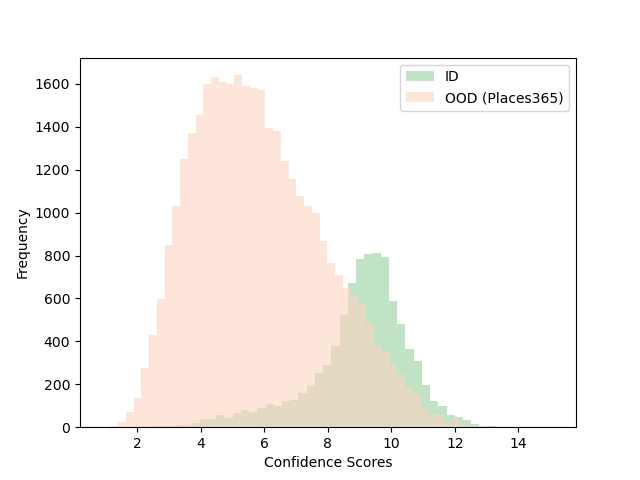}}
		\subfigure{\includegraphics[width=4cm]{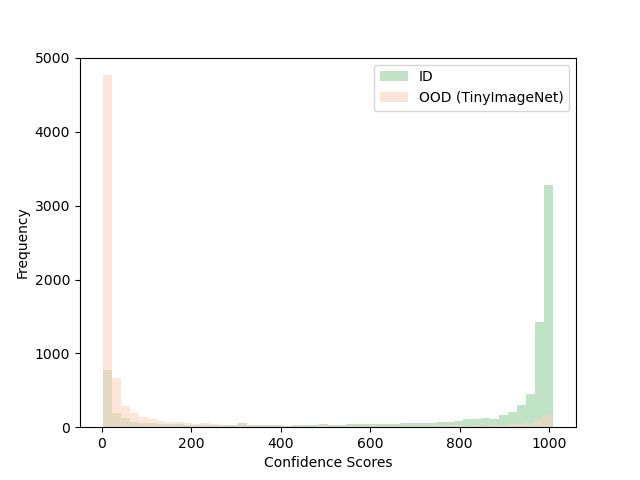}}
		\subfigure{\includegraphics[width=4cm]{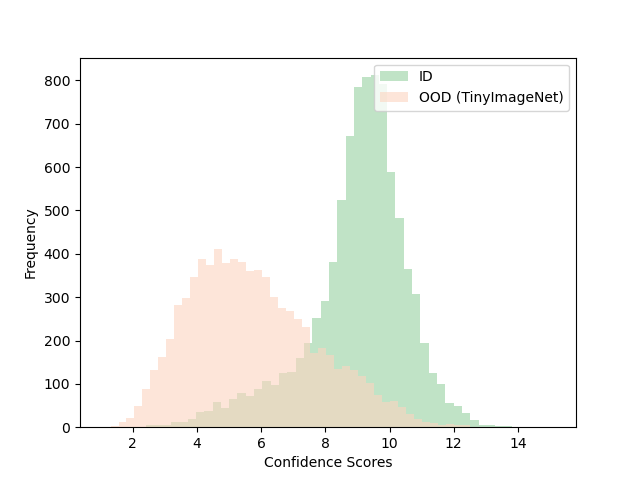}}
	\end{center}\vspace*{0pt}\caption{This figure displays the difference between max logits and class relevance learning. We train the ResNet18 model on the CIFAR10 dataset and test it on the TinyImageNet \cite{le2015tiny} and Places365 dataset. The left column shows the confidence score distribution of CRL and right column shows the distribution of Maxlogits. }
	\label{fig:compare_vis}
\end{figure}

\textbf{Visualization of confidences scores}
The elucidation of distinctions between max logits and class relevance learning is explicated in Figure~\ref{fig:compare_vis}. The ResNet18 architecture is employed as the foundational neural network, undergoing training on the CIFAR10 dataset. Subsequently, an evaluation is conducted on the TinyImageNet dataset \cite{le2015tiny} and the Places365 dataset, adhering to the protocol established by Yang et al. \cite{yang2022openood}. In the leftmost column of the figure, the confidence score distribution of the proposed Class Relevance Learning (CRL) method is portrayed, while the rightmost column illustrates the Maxlogits approach.

Evidently, the CRL method manifests a superior confidence distribution for out-of-distribution (OOD) detection. A discernible distinction arises, wherein the majority of OOD samples are conspicuously distinguished. Furthermore, the demarcation between in-distribution (ID) samples and OOD samples is more distinctly discerned. In contrast, the Maxlogits approach exhibits a higher proportion of samples that reside in the ambiguous region between the ID and OOD boundaries, rendering them challenging to differentiate.

\section{Conclusion}\label{sec:conclusion}
In this paper, we propose a class relevance learning for OOD detection. Unlike previous methods, which only exploit the single logits/softmax probability, we first build up the class relevance concept by statistically analyzing the inter class relationship and constructing a class relevance matrix. During the test stage, the logits and class relevance matrix are utilized for OOD score estimation. The result will serve as an OOD score to distinguish which sample is OOD. Experiment results on diverse OOD benchmarks show CRL has superior performance than previous state-of-the-art methods.


\newpage
\small
\bibliographystyle{IEEEbib}
\bibliography{ood}
\end{document}